\documentclass[10pt,twocolumn,letterpaper]{article}

\usepackage[pagenumbers]{cvpr} 

\usepackage{graphicx}
\usepackage{amsmath}
\usepackage{amssymb}

\usepackage{booktabs}

\usepackage[capitalize]{cleveref}
\crefname{section}{Sec.}{Secs.}
\Crefname{section}{Section}{Sections}
\Crefname{table}{Table}{Tables}
\crefname{table}{Tab.}{Tabs.}

\def\cvprPaperID{*****} 
\def\confName{CVPR}
\def\confYear{2023}

\begin{document}

\title{Enhancing Multi-Camera People Tracking with \\Anchor-Guided Clustering and Spatio-Temporal Consistency ID Re-Assignment}

\author{Hsiang-Wei Huang$^{1*}$\ \ Cheng-Yen Yang$^{1*}$\ \  Zhongyu Jiang$^{1}$\ \ Pyong-Kun Kim$^{2}$\\ Kyoungoh Lee$^{2}$\ \ Kwangju Kim$^{2}$\ \  Samartha Ramkumar$^{1}$\ \  Chaitanya Mullapudi$^{1}$\\  In-Su Jang$^{2}$\ \  Chung-I Huang$^{3}$ \ \ Jenq-Neng Hwang$^{1}$ \vspace{0.5em}\\
$^{1}$ Information Processing Lab, University of Washington, USA\\
$^{2}$ Electronics and Telecommunications Research Institute, South Korea \\
$^{3}$ National Center for High-Performance Computing, Taiwan\\
}

\maketitle

\begin{abstract}
Multi-camera multiple people tracking has become an increasingly important area of research due to the growing demand for accurate and efficient indoor people tracking systems, particularly in settings such as retail, healthcare centers, and transit hubs. We proposed a novel multi-camera multiple people tracking method that uses anchor-guided clustering for cross-camera re-identification and spatio-temporal consistency for geometry-based cross-camera ID reassigning. Our approach aims to improve the accuracy of tracking by identifying key features that are unique to every individual and utilizing the overlap of views between cameras to predict accurate trajectories without needing the actual camera parameters. The method has demonstrated robustness and effectiveness in handling both synthetic and real-world data. The proposed method is evaluated on CVPR AI City Challenge 2023 dataset, achieving IDF1 of 95.36\% with the first-place ranking in the challenge. The code is available at: \url{https://github.com/ipl-uw/AIC23_Track1_UWIPL_ETRI}.

\end{abstract}


\section{Introduction}
\label{sec:intro}


Multi-people tracking, which involves detecting and monitoring human movement, has become an essential tool in various industries. Such tracking utilizes techniques like sensors, cameras, and deep learning algorithms to track people's positions, motions, and directions with time. It plays a critical role in ensuring security surveillance and bunisness analytics as well as works with closed-circuit television (CCTV) to prevent accidents. 

\begin{figure}[!t]
\includegraphics[width=\linewidth]{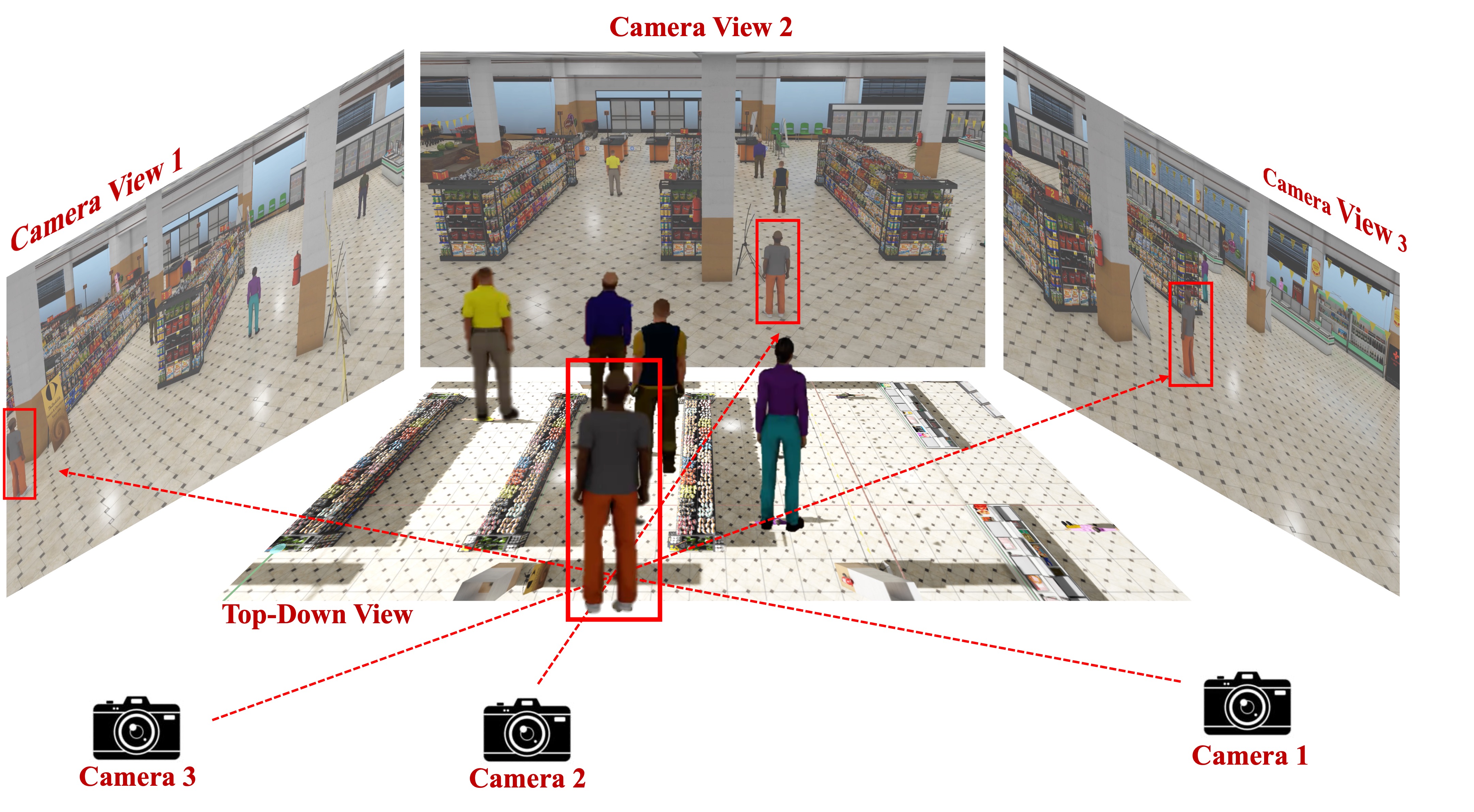}
\centering
\caption{Illustration of \textbf{M}ulti-\textbf{C}amera \textbf{P}eople \textbf{T}racking (\textbf{MCPT}). The task involves detecting and tracking the same individuals across multiple cameras. The goal is to maintain the identity of each individual and their trajectory across different views, while dealing with challenges such as occlusion and camera viewpoint variations. }
\end{figure}

\let\thefootnote\relax\footnotetext{* indicates equal contributions.}


The demand for indoor people tracking has increased in recent years. Besides tracking people's movements, indoor environments necessitate advanced technology to detect and monitor people's activities. In the healthcare sector, it is vital for tracking patients and staff, equipment and inventory, and optimizing workflows \cite{yao2012adoption}. Similarly, the retail industry can enhance the shopping experience by analyzing customer behavior, and security and safety can leverage this technology to detect and respond to emergencies. Moreover, the recent COVID-19 pandemic has underscored the need for quarantine measures such as social distancing \cite{punn2020monitoring}.

However, due to the privacy issue, the data are limited for researching deep learning based people tracking methods.
Therefore, researchers are exploring the use of synthetic imagery as an alternative\cite{tremblay2018training, wang2021real, kortylewski2018training}. Synthetic imagery mimics real-world footage and can be used as training data for machine learning models to create large training datasets in a cost-effective manner. This approach also benefits consistency and predictability, enabling customization for specific user needs, such as class balance considerations. Moreover, it can enhance the generalization capability of models that are challenging to train with limited real-world data and address privacy concerns. However, since synthetic data may not capture the full complexity and variability of real-world data, models trained on this data could be biased or inaccurate.

The AI City Challenge recently released data on indoor people tracking using synthetic videos under mutli-camera settings. The dataset focuses on multi-camera cross-view scenarios. So, we develop a multi-camera people tracking method (MCPT) composed of three main components: single-camera tracking, anchor-guided clustering for multi-camera re-identification, and 3D-based spatio-temporal consistency ID re-assignment for post-processing. Our proposed method outperformed all others in AI City Challenge Track 1, resulting in the best performance. Therefore, we assert three main contributions in this paper:

\begin{itemize}
\setlength{\itemsep}{0pt}
\setlength{\parsep}{0pt}
\setlength{\parskip}{0pt}
    \item We present a robust anchor-guided clustering method for multi-camera people tracking and re-identification.
    \item We leverage the spatio-temporal consistency of each track for post-processing enabled by self-camera calibrations. which can significantly improve the tracking accuracy of people with similar appearances.
    \item Achieved the best performance with an IDF1 of $95.36$, in the 2023 AI City Challenge Track 1 on the public testing set which consists of data from real and synthetic multi-camera settings.
\end{itemize}


The subsequent sections of the paper are structured as follows: Section \ref{sec:rw} provides an overview of related works. Section \ref{sec:method} describes the proposed method. Section \ref{sec:implementation} presents the results of the detailed implementation and experiment results. Lastly, Section \ref{sec:conclusion} provides the discussion and conclusions drawn from the study.

\section{Related Works}
\label{sec:rw}

\subsection{Re-Identification}


Since the the advent of deep learning technology, CNN features have been used dominantly \cite{ye2021deep}, studies on person re-identification have been conducted from three main categories: network structure, loss definition, and sampling method.

\noindent{\textbf{Network Structure.}} \cite{ahmed2015improved} extracted features from each of the two images and computed relationships between them to train a model. \cite{dai2019batch} reviewed prior research on feature drop and localizing different body parts as their proposed model included a global branch for encoding global salient representations and a feature-dropping branch for randomly dropping the same region of all input maps in a batch. \cite{zhou2021learning} proposed a CNN architecture that leverages multiple scales with different receptive field sizes and dynamically fused them using channel-wise adaptive aggregation. They demonstrated that their system was significantly smaller than previous models by utilizing factorized convolutions in the building blocks. \cite{lee2022reet} highlighted the tokens generated by the region level and introduced a region-based feature pooling method for obtaining more granular areas of interest.

\noindent{\textbf{Loss Definition.}} \cite{schroff2015facenet} introduced a new loss called triplet loss, which has inspired numerous subsequent studies in re-id and metric learning. \cite{hermans2017defense} demonstrated that a variant of triplet loss outperformed other losses, which contradicted the prevailing belief that triplet loss was inferior to surrogate loss functions.

\noindent{\textbf{Sampling Method.}} Early work typically used random sampling\cite{bell2015learning, chopra2005learning}. \cite{schroff2015facenet} proposed semi-hard negative mining, which required a large batch size. \cite{simo2015discriminative} utilized hard negative mining in a siamese network. \cite{wu2017sampling} focused on selecting more informative and stable examples than traditional approaches by employing a margin-based loss.


\begin{figure*}[t]
\includegraphics[width=\textwidth]{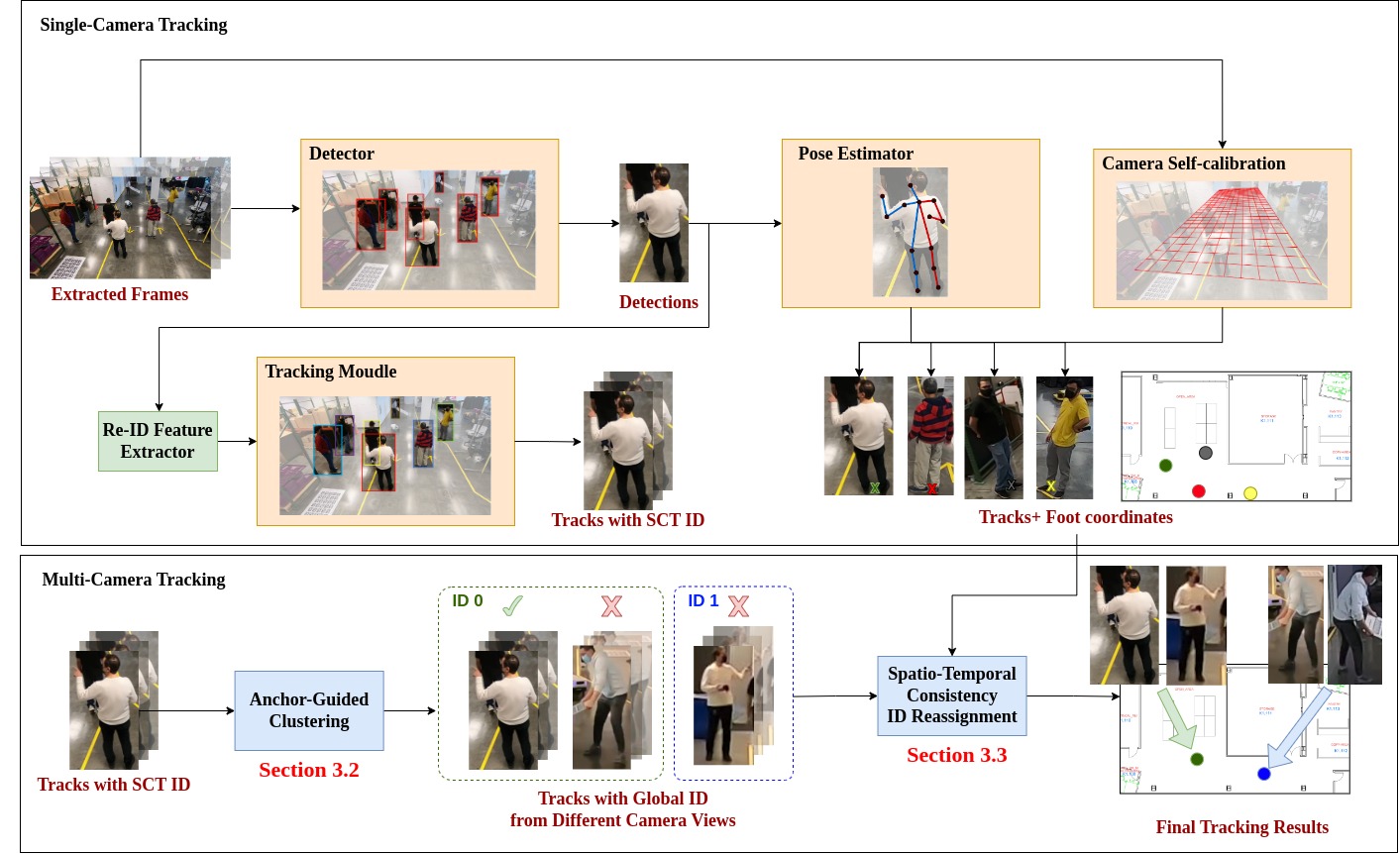}
\centering
\caption{The overall pipeline of our Multi-Camera Multi-people Tracking (MCMT) framework. (1) \textbf{Single-Camera Tracking} is first conducted with a standard tracking-by-detection scheme, where the extracted frames are fed into a detector and a feature extractor to get the detections and re-id features for the preliminary tracking.  (2) Then \textbf{Anchor-Guided Clustering} will assign a global ID as well as fixing the ID switches for each trajectory. (3) Lastly, \textbf{Spatio-Temporal Consistency ID Reassignment} will utilize the 2D human pose with camera self-calibration to reproject on the map for final post-processing.}
\end{figure*}

\subsection{Monocular Multi-Object Tracking}

The field of Monocular Multi-Object Tracking has advanced significantly since the advent of deep learning technology, with studies typically following the tracking-by-detection paradigm \cite{andriluka2008people, breitenstein2009robust, kalal2011tracking, yu2016poi, bewley2016simple, wojke2017simple, aharon2022bot}. This involves detecting the location of objects to be tracked in each frame and associating them based on the similarity of their real and predicted locations, which is calculated using motion information to produce a trajectory. Early studies \cite{bochinski2017high} explored the use of the Intersection of Union (IoU) metric to improve tracking speed. Another work \cite{bewley2016simple} focused on improving tracking performance with simple motion prediction using a Kalman filter to predict object location in the next frame. While this approach is fast and effective in simple environments, it has limitations in complex scenarios with a high number of objects to track. Other studies, such as \cite{wojke2017simple}, focused on using appearance information to match objects, by extracting feature vectors from detected objects and comparing them across frames.

Many studies \cite{bergmann2019tracking, voigtlaender2019mots, xu2020segment, wang2020towards, du2023strongsort, zhang2021fairmot, zhang2022bytetrack, huang2023observation, yang2023multiobject} have proposed various approaches to enhance Multi-Object Tracking (MOT) performance with low computational cost. \cite{bergmann2019tracking} addressed the issue of the lack of a tracking dataset compared to the detection dataset. Meanwhile, \cite{voigtlaender2019mots} thought that bounding box level tracking is saturating and introduced pixel-level tracking as a way to improve performance. Other studies, such as \cite{xu2020segment}, focused on learning instance embeddings using 2D point cloud representations to avoid background features. \cite{wang2020towards} proposed a real-time Joint Detection and Embedding (JDE) system for MOT, which outperformed the Separate Detection and Embedding (SDE) system used in previous studies. \cite{zhang2021fairmot} treated MOT as a multi-task learning problem of object detection and association and presented detailed designs to avoid competition between these tasks. \cite{du2023strongsort} demonstrated that simple designs could perform well with a few additional tricks. \cite{zhang2022bytetrack} proposed a method that associated almost every detection box, rather than just the high score ones. \cite{cao2022observation} computed a virtual trajectory over the occlusion period based on object observations. Finally, \cite{aharon2022bot} proposed some bag of tricks to achieve high performance on public MOT datasets.

\subsection{Multi-Camera Multi-Object Tracking}

Following in the development of single-camera multi-object tracking, Multi-camera multi-object tracking has been studied actively. Previous studies were based on the graph-based approaches to associate across frames and cameras \cite{chen2016equalized, wen2017multi, hofmann2013hypergraphs, he2020multi}. As is the case with single cameras, the deep feature was soon introduced in the multi-cameras \cite{ristani2018features, hsu2020traffic, specker2021occlusion}. \cite{ristani2018features} proposed an adaptive weight loss and hard-identity mining scheme for learning better features.  \cite{hsu2020traffic, huang2023multi} proposed the trajectory-based camera link model including deep feature re-identification. They utilized the TrackletNet Tracker (TNT) to generate the moving trajectories and the camera link model to constrain the order by the spatial and temporal information. \cite{yang2023unified} proposed a unified framework that can effectively adopt monocular 2D bounding boxes and 2D poses jointly to produce robust 3D trajectories to track across mutli-camera with overlapping views. \cite{nguyen2022lmgp} proposed multi-camera multiple object tracking approach based on a spatial-temporal lifted multicut formulation utilizing 3D geometry projection.

The release of many public datasets has driven progress in this field. \cite{naphade20182018, tang2019cityflow, Naphade22AIC22} released a city-scale traffic camera dataset consisting of more than 3 hours of HD videos. \cite{han2023mmptrack} proposed a novel method to construct a large-scale multi-camera tracking dataset called MMPTrack to alleviate the occlusion issue. They utilized depth and RGB cameras to build 3D tracking results and projected them to create 2D tracking results. This helped to build a reliable benchmark for multi-camera multi-object tracking systems in cluttered and crowded environments.

\begin{figure*}[!htb]
\includegraphics[width=0.96\textwidth]{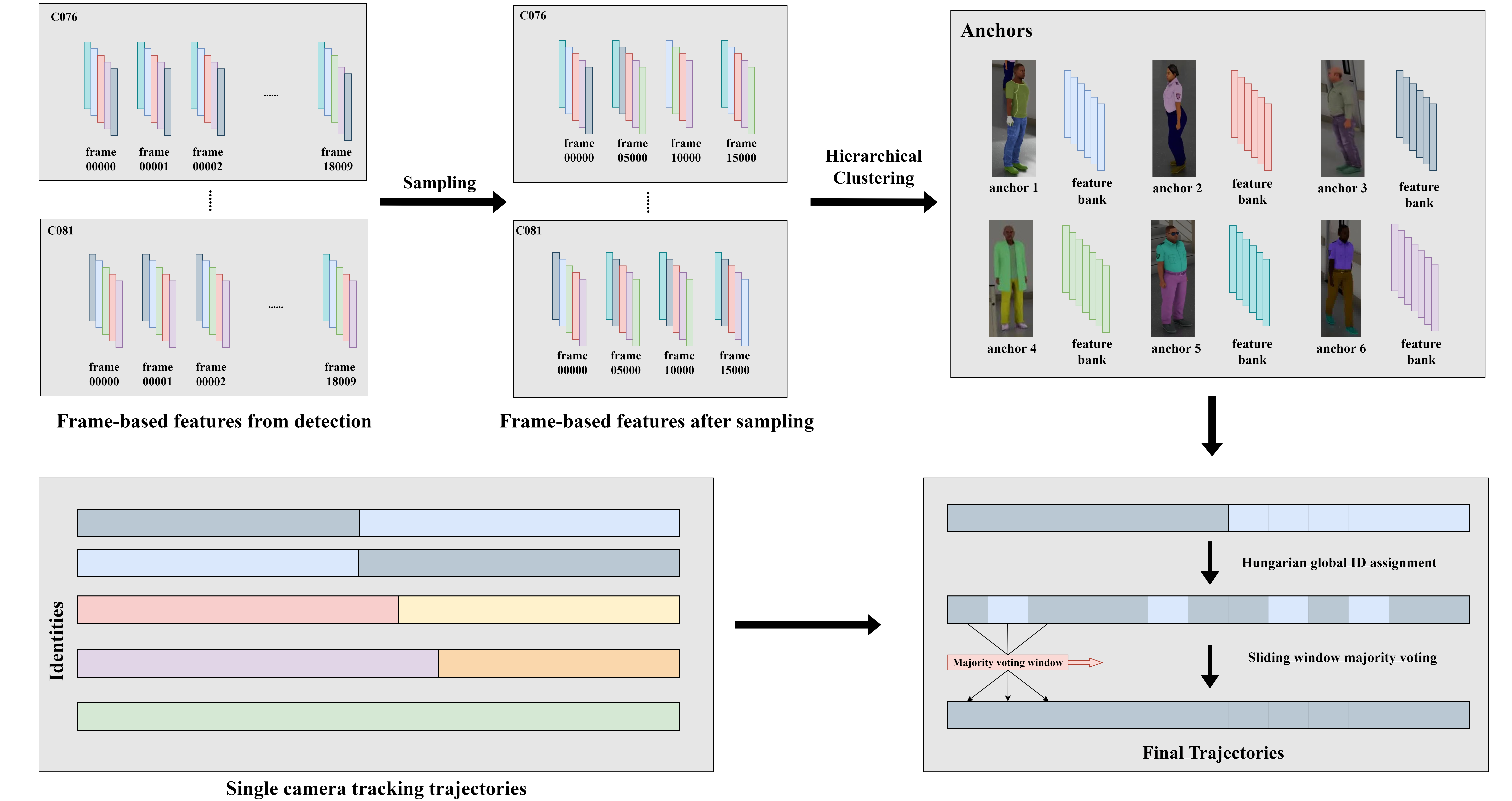}
\centering
\caption{The process for anchor-guided multi-camera people tracking involves periodically sampling frame-based appearance features from each camera, and performing hierarchical clustering to obtain anchors with corresponding feature banks. Single camera tracking is then performed to obtain preliminary trajectories with ID switches denoted in different colors in the same row. Each detection from the preliminary trajectories are further assigned a global ID by the anchor using the Hungarian algorithm. Finally, sliding window majority vote is performed to obtain the final trajectories.}
\end{figure*}

\section{Methods}
\label{sec:method}

\subsection{Single-Camera Tracking}
Single-camera tracking algorithms generally follow the tracking-by-detection paradigm, which involves using an independent detector on the input image and then an association algorithm to link bounding boxes across frames. To incorporate appearance features into the tracking process, we employ the Re-ID version of BoT-SORT \cite{aharon2022bot} as a single-camera tracking algorithm to obtain preliminary object trajectories. BoT-SORT is designed to leverage both object motion and appearance for the association, which has led to achieving state-of-the-art tracking performance on several multi-object tracking benchmarks.


\subsection{Anchor-Guided Clustering}

Upon completion of single-camera tracking, preliminary trajectories are often obtained, which may suffer from numerous ID switches due to occlusion, unaligned object IDs of the same identity between different cameras in the same scenario, and a lack of a re-entry handling method in the camera field of view. The absence of such a method makes it difficult to assign a unique object ID to identities that re-enter the camera field of view after exiting for a certain time.

To address all three of these problems simultaneously, we propose an effective method called anchor-guided clustering and global ID assignment. Our method elegantly enables the assignment of the same unique object ID to re-entering identities, and it can handle occlusion-induced ID switches and unaligned object IDs of the same identity between different cameras in the same scenario.

The proposed method involves periodic sampling of detection appearance features for a certain number of frames from each camera in the same scene. After the sampling process is completed, hierarchical clustering is performed to obtain several anchors, where each anchor contains multiple features that represent the same identity's appearance feature under different detection sizes, lighting conditions, and rotation angles. Each anchor has a unique ID that will be used as the identity's global ID in multi-camera tracking.

Subsequently, the detections in the same frame $t$ and the anchors will perform the Hungarian algorithm with cost described in the following formula:

\begin{equation}
\text{cost}(d_{i,t}, a_j) = 1 - \frac{1}{k}\sum_{l=1}^{k}\frac{d_{i,t} \cdot a_{j,l}}{\left| d_{i,t} \right| \left| a_{j,l} \right|}
\end{equation}

We use the cosine distance between the detection's appearance feature $d_{i,t}$ (where $t$ is the frame ID of the detection) and each appearance feature vector $a_{j,k}$ in the anchor $a_j$ as the cost for global ID assignment. Specifically, we compute the average of the cosine distances between $d_{i,t}$ and all $a_{j,k}$ to obtain the final cost. Note that each anchor contains multiple appearance feature vectors, denoted as $a_{j,l}$, and the number of these vectors are denoted as $k$.

After performing the Hungarian algorithm, each single-camera tracking trajectory obtains a global ID list with the same length as the original trajectory. To assign the final global ID, we use a sliding window majority voting approach. This method effectively fixes ID switches in single-camera tracking, assigns global IDs robustly by considering multiple frames, and correctly re-identifies individuals who re-enter the camera field of view by assigning them the same global ID.

\subsection{Spatio-Temporal Consistency ID Reassignment}

Assuming that the multi-view videos are synchronized and overlapped, it is expected that a person's trajectories will exhibit both \textbf{spatial} and \textbf{temporal consistency} in terms of \textbf{position} and \textbf{motion} across all views. Therefore, by relying on such cross-view consistency, it becomes possible to match the 2D tracklets under different views to the same person and further re-assigned the incorrect global IDs in the previous tracking stage either due to similar appearance or heavy occlusion.

Given the tracking results after the anchor-guided clustering ID assignment, where $X^{k}_{t,id} \in \mathbb{R}^4$ representing the 2D information of each detection $(x,y,w,h)$ under $k$-th camera-view at frame $t$. The function $F_{g}$ takes any detection $X^{k}_{t,id}$ as input and outputs 2D coordinates in image space representing the ground-plane location on which each target is located whenever both the left and right ankle keypoints $(x_{la}, y_{la})$ and $(x_{ra}, y_{ra})$ are available:

\begin{equation}
    F_{g}(X^{k}_{t,id}) = \begin{cases}
        (\frac{(x_{la}+x_{ra})}{2}, \frac{(y_{la}+y_{ra})}{2}) & \text{if } c_{la},c_{ra} \geq \tau_{pose}\\
        (x+w/2, y+h)  & \text{otherwise}
    \end{cases} 
\end{equation}

\noindent where $c_{la}$ and $c_{ra}$ are the confidence scores of the keypoints predicted by the 2D pose estimator. The $\tau_{pose}$ is the thresholding that controls whether we decided the top-down location of each target by pose-based analysis or simply compute from the bounding box's information. 

Then, given the homography matrix $H^{k} \in \mathbb{R}^{3\times3}$ of $k$-th camera-view obtained via camera self-calibration, we can obtained the top-down coordinate of any detection $X^{cam}_{t}$ by reprojecting the ground-plane coordinates of each detections in image space using:

\begin{equation}
    F_{3D}(X^{k}_{t,id}) = H^{k} \cdot F_{g}(X^{k}_{t,id})^T,
\end{equation}

\noindent we hereby used the notation $\hat X^{k}_{t,id} \in \mathbb{R}^6$ representing the 2D information of each detection $(x,y,w,h)$ and 3D information $(x_{3D}, y_{3D})$ under $k$-th camera-view at frame $t$.

The spatial consistency in our work refers to the level of agreement of the top-down location of each ID from all of the camera views. With $\hat X^{k}_{t,id}$ representing the 3D information $(x_{3D}, y_{3D})$ under $k$-th camera-view at frame $t$, the spatial consistency across multi-view is defined as:

\begin{equation}
    D_{spatial}(\hat X^{k}_{t, id}, t, id) = \frac{1}{N}\sum_{l \neq k} || \hat X^{l}_{t, id} - \hat X^{k}_{t, id} ||^{2},
\end{equation}

\noindent as it is worth mentioning that we will exclude the outliers identified by the function $O(\cdot)$ prior to computing the average coordinates. These outliers are typically detections with similar appearances that are likely to be misclassified as a different identity in our previous single-camera tracking or anchor-guided clustering. Finally, we use a self-defined confidence score as a threshold to determine which detections we will reassign identities to:

\begin{equation}
    conf_{i\rightarrow j}(\hat X^{k}_{t, i}) = 1 - \frac{D_{spatial}(\hat X^{k}_{t, i}, t, j)}{D_{spatial}(\hat X^{k}_{t, i}, t, i)}
\end{equation}

In addition, if we regard any sudden changes in the locations of tracks as irregular, we can enhance their consistency over time by employing a straightforward method of performing a weighted summation of track locations within a sliding window. This generates a smoothed location estimate that captures the general movement of the track over time. This method can be highly beneficial in scenarios where the data contains noise or missing information that may result in abrupt changes in track locations, which are not due to actual movements but rather measurement errors. We can establish temporal consistency by replacing the initial average coordinate calculation with a weighted variant. This involves assigning weights to the coordinates based on their relevance to the sliding time window, and computing the weighted average instead.

\section{Implementation Details}
\label{sec:implementation}

\subsection{Dataset}

The Multi-camera People Tracking dataset consists of multiple camera feeds captured in various real-world and synthetic settings. The real-world data were collected from a warehouse while the large-scale synthetic data were synthesized using the NVIDIA Omniverse Platform across six different indoor scenes. The videos are in high-resolution 1080p feeds at 30 frames per second with tracking annotations across camera views. However, it is important to note that there are 10 and 5 synthetic sets in the training and validation data respectively while there are 6 synthetic sets plus an additional real set in the testing data. Our experiments exclusively used the data from the dataset and did not incorporate any external data, whether real or synthetic.

Furthermore, for the real data in the testing set, we only employed pre-trained detector, 2D pose estimator, and re-id feature extractor to demonstrate the effectiveness and robustness of our proposed multi-camera people tracking method.

\begin{table}[t]
\small
\centering
\begin{tabular}{lccc}
\hline
Dataset Split & \# of Scenes (Cams) & \# of Frames & \# of Dets \\ \hline
Training (syn)     & 10 (58)                & 1,065,602    & 4,375,736        \\
Validation (syn)    & 5 (28)                  & 504,252      & 1,950,917        \\
Testing (real)   & 1 (7)                &  388,671     & -                \\
Testing (syn)    & 6 (36)               &  648,360     & -                \\ \hline
\end{tabular}
\caption{Basic information of the Multi-camera People Tracking dataset presented in 2023 AI City Challenge Track 1. The annotations of the testing data are and will most likely remain private therefore no accurate number of total detections can be provided.}
\end{table}

\subsection{Evaluation Metrics}

We adopt the mean Average Precision (mAP) for detection-related tasks while using the Rank-1 accuracy for Re-ID tasks. As for the multi-camera tracking, we will use the IDF1 score, which measures the ratio of correctly identified detections over the average number of ground-truth and computed detections. The challenge submission platform also provided with other MOT-related evaluation measures, such as IDF1, IDP, IDR, precision (detection), and recall (detection). Other MOTChallenge evaluation measures, such as MOTA, MOTP, MT, and FAR will not only be used as self-evaluation metrics.

\subsection{Camera Calibration}

The WILDTRACK and MMP-Track datasets provide camera calibration files with the pinhole camera model with both extrinsic and intrinsic parameters for each camera given. However, in Challenge Track 1: Multi-Camera People Tracking (MCPT), the calibration is not provided with the dataset but the top-down view map is available for each subset.

We choose correspondences between the camera-view frames and the top-down view map to compute the homography matrices $H$ for each camera in order to project the ground-plane location for each target from the 2D image space to the 3D world space. We adopt the semi-automatic camera calibration based on the Perspective-n-Point method to compute the homography matrix for each camera. For each camera view, we manually select 6 to 12 pairs of points as input, using the approaches including (1) a Least-Squares method using all the points, (2) a RANSAC-based robust method, (3) a Least-Median-of-Squares method or (4) a PROSAC-based robust method.

\subsection{Detector}

\noindent{\textbf{Synthetic Data.}} With the high frame rate at 30, the training and validation sets consist of 1M and 500k high-resolution frames each from 15 sequences with a total of 6.3M detections. In order to maintain a balance between the long training elapsed time and the detection performance, we eventually decide to conduct our first-stage of pretraining under a sampling rate of 20 using all the scenes and second-stage of fine-tuning under a sampling rate of 15 for specific scene.

We use YOLOv7 as our backbone and COCO-pretrained weights from \cite{mmyolo2022} for initialization. Our first-stage pretrained model were trained for 60 epochs with a batch size of 8 and an initial learning rate of 0.0025. Our second-stage scene-specific fine-tuned models were trained for 10 epochs with a batch size of 8 and an initial learning rate of 0.00025. 

\noindent{\textbf{Real Data.}} The only real world data in the dataset is sequence \textit{S001} from test split. Although supervised and unsupervised domain adaptation methods on object detection \cite{da_od_survey, uda_od_survey} show promising results on label-scarce target datasets. The lack of corresponding real data in training or validation split make it difficult for us to evaluate the performance of cross-domain human detection. Therefore for the purpose of challenge, we directly employ the public available pretrained YOLOX\_x model from \cite{zhang2022bytetrack} train on CrowdHuman, MOT17, Cityperson and ETHZ.

This allows us to have a benchmark performance for cross-domain human detection on the dataset, but further research and experimentation would be necessary to improve the performance of the models on the dataset. It is also important to note that the lack of real-world data in the dataset may limit its applicability to certain real-world use cases, and therefore, collecting more diverse and representative data would be necessary for the dataset to be more widely applicable.

\noindent{\textbf{2D Pose Estimation.}} Since there are no 2D pose annotations in the dataset, we directly impose the top-down human pose estimation method, HigherHRNet, using pre-trained weights from \cite{han2023mmptrack}. The inputs are cropped out based on the bounding boxes predict from the YOLO detector, then 17 keypoints are estimated under COCO format.

\begin{figure*}[!htb]
\includegraphics[width=0.95\textwidth]{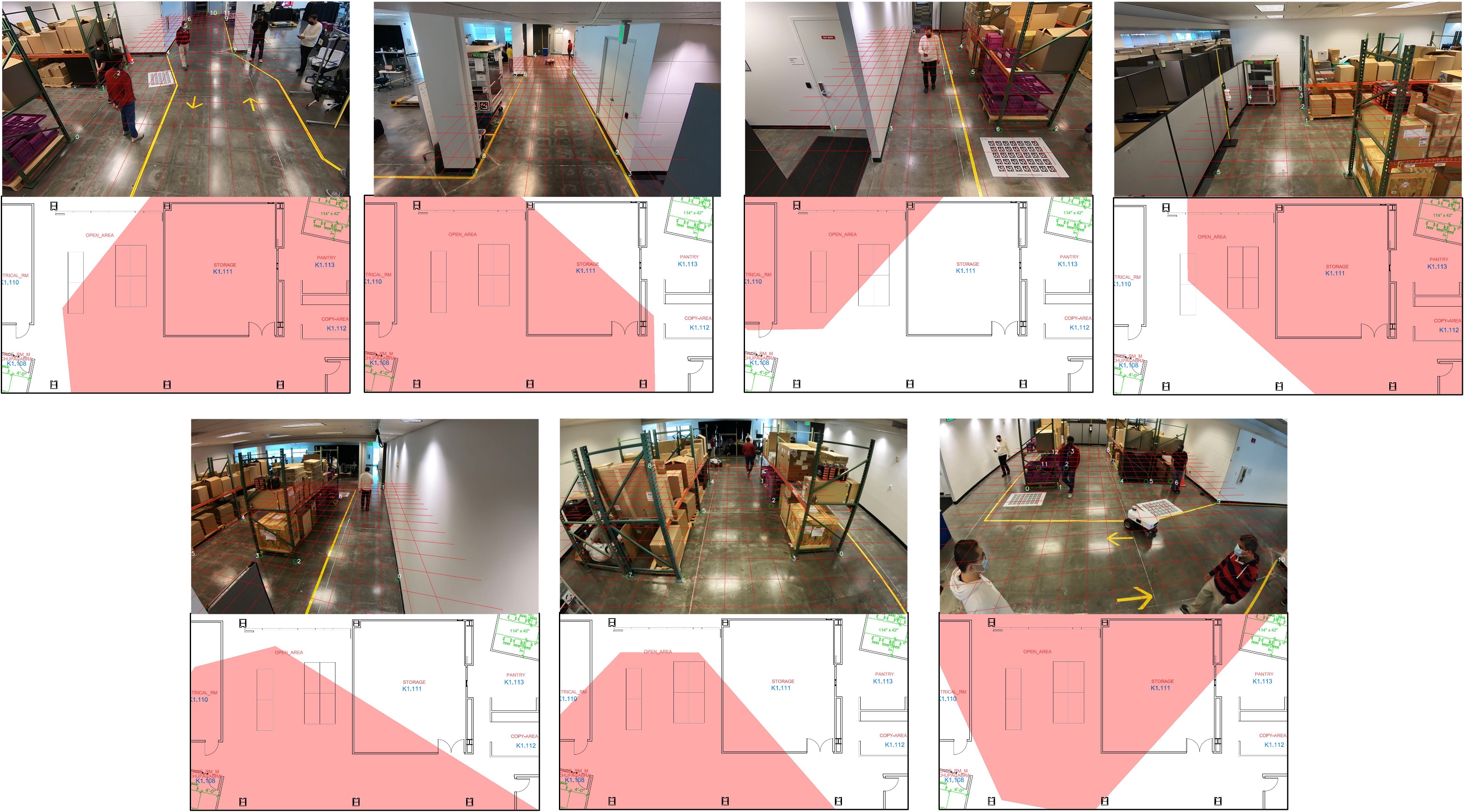}
\centering
\caption{Top: Self-camera calibration visualizing the estimated ground plane (red grid lines) for different camera views for the testing sequence \textit{S001}. Bottom: Field of view for different cameras projected on the top-down map for the testing sequence \textit{S001} .}
\end{figure*}

\subsection{Re-ID Model}
In our multi-camera people tracking system, we have chosen OSNet as the person re-identification (ReID) model. The OSNet architecture has shown to be effective in person ReID tasks using the unified aggregation gate to fuse the features from different scales and has achieved state-of-the-art performance on several benchmark datasets. 

\noindent{\textbf{Synthetic Data.}} 
The ReID training data is sampled from the training and validation set of the 2023 AI City Challenge Track1 dataset. We random sample each trajectory and divided the samples into training, testing, and query sets. The ReID dataset used in our training process contains 56,181 training images, 17,027 testing images, and 2,846 query images from different cameras and scenes. The OSNet is trained for 60 epochs, Adam optimizer, and 0.0003 learning rate with data augmentation like random flip. The final model achieves 97.9\% Rank-1 accuracy and 97.6\% mAP on the sampled testing set.


\begin{table}[t]
  \begin{center}
    {\small{
\begin{tabular}{ll}
\toprule
Model & Re-ID Dataset\\ 
\midrule
OSNet & Market1501\\
OSNet & MSMT17\\
OSNet & Market1501 + CUHK03 + MSMT17\\
OSNet-IBN & Market1501 + CUHK03 + MSMT17\\
OSNet-AIN & Market1501 + CUHK03 + MSMT17\\
\bottomrule
\end{tabular}
}}
\end{center}
\vspace{-1em}
\caption{Pre-trained models used for re-id on the real scene.}
\end{table}

\noindent{\textbf{Real Data.}} 
Due to the lack of real-world data in the training set of the challenge, several pre-trained models on other human Re-ID datasets are used for the multi-camera tracking in the real-world scenario in the testing set.
Three different model architectures are used, including OSNet, OSNet-IBN, and OSNet-AIN. A total of five models pre-trained on different dataset combinations are used, the model architectures and the pre-trained dataset used can be found in the following table. The features extracted from these models are directly concatenated together for the use of single-camera tracking and multi-camera tracking.

\subsection{Tracking}

\noindent{\textbf{Single Camera Tracking.}} For filtering the detection results, the synthetic scenario tracking has the high score threshold of 0.6 and low score threshold of 0.1. For the real world scenario, the high score threshold is 0.6 while the low score is carefully fine-tuned in each camera. All the other parameters used in the single camera tracking is the default parameters of BoT-SORT. Since the camera is stationary, the camera motion compensation part is removed to reduce computational cost.


\begin{figure*}[t]
\includegraphics[width=0.93\textwidth]{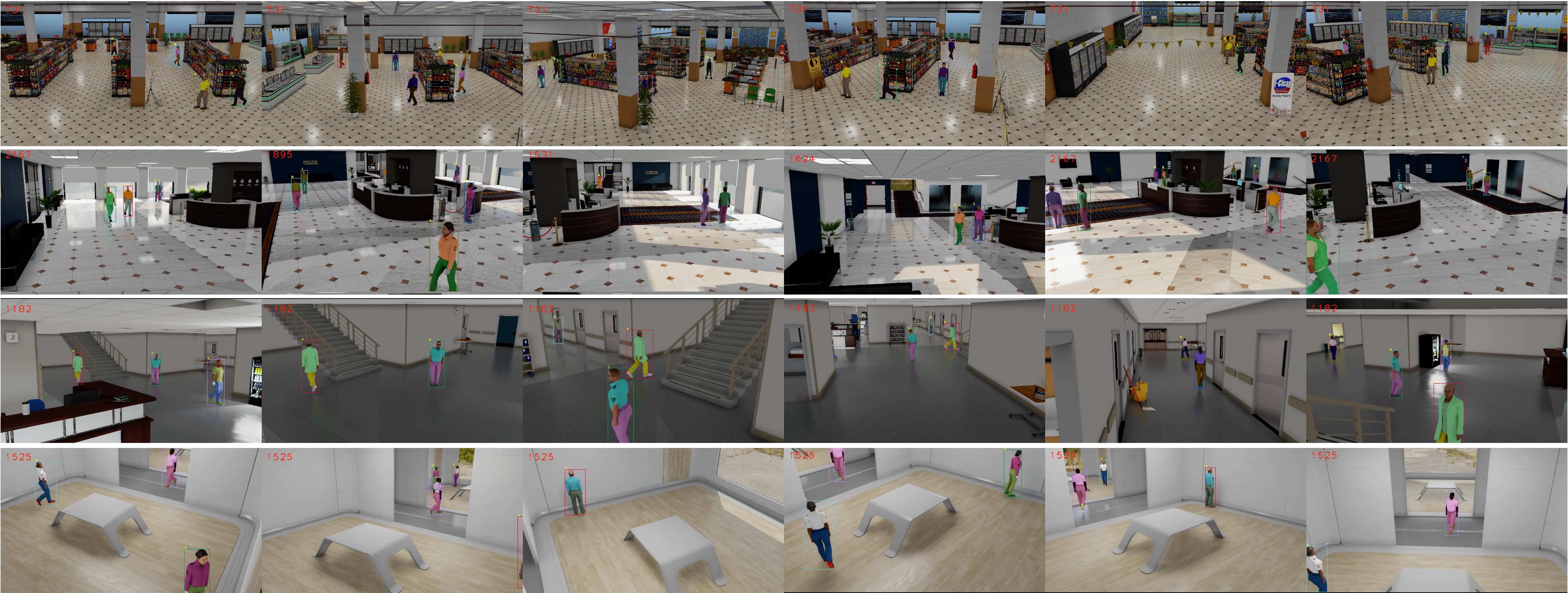}
\centering
\caption{Visualization of our final tracking results on the testing synthetic dataset.}
\end{figure*}

\noindent{\textbf{Mutlil-Camera Tracking.}} There are several parameters in the multi-camera tracking system, including the hierarchical clustering threshold and the length of majority vote sliding windows. The hierarchical clustering threshold is carefully fine-tuned to make sure the anchor cluster results are accurate. The length of sliding windows should be big enough to achieve robustness in the voting process, while it can not be too big so that the ID switch can not be fixed immediately after ID switch happened in the single camera tracking. In our final system, we use the majority vote length of 15 to achieve balance between ID assignment robustness and the ability for ID correction in single camera tracking. Finally, linear interpolation is performed to all the tracking results before submission.

\subsection{Results on AI City Challenge}
Several methods with different models and post-processing are evaluated on the public test set of the 2023 AI City Challenge Track 1 \cite{Naphade23AIC23} as shown in Table \ref{table:ablation}. Our baseline method is the anchor-guided clustering multi-camera people tracking approach without any spatial-temporal re-assignment, which achieves an IDF1 score of 89.57\%. We then conduct experiments with three variations of the method: using the \textbf{F}ine-\textbf{T}uned detector models for each specific scene (\textbf{FT}), introducing \textbf{S}patio-\textbf{T}emporal \textbf{C}onsistency \textbf{R}e-\textbf{A}ssignment (\textbf{STCRA}) into the framework, and iterative refinement of the latter technique, named \textbf{i}terative spatio-temporal consistency re-assignment (\textbf{i-STCRA}). For i-STCRA, which is our final submission, we use a $k=3$ with an ascending confidence score thresholding and a descending outlier thresholding to ensure that the re-assignments are stricter after each iteration.

Our experiments demonstrate that each variation leads to improved performance, with the best result achieved by the method with spatio-temporal consistency iterative re-assignment, obtaining an IDF1 score of 95.36, IDP score of 95.83, and IDR score of 94.88 ranking the first-place among 27 teams as shown in Table \ref{table:ranking}.



\begin{table}[t]
\centering
\small
\begin{tabular}{lccccc}
\hline
Method         & IDF1  & IDP   & IDR   & Precision & Recall \\ \hline
Baseline      & 89.57 & 91.89 & 87.36 & 92.79     & 88.21  \\
+ FT  & 92.98 & 92.01 & 93.97 & 92.83     & 94.81  \\
+ FT + STCRA  & 93.62 & 92.90 & 94.35 & 93.61     & 95.08  \\
+ FT + i-STCRA & \textbf{95.36} & \textbf{95.83} & \textbf{94.88} & \textbf{96.44}     & \textbf{95.49}  \\ \hline
\end{tabular}
\vspace{-.5em}
\caption{The experimental results on the public test set of Track 1.}
\vspace{-1.em}
\label{table:ablation}
\end{table}

\begin{table}[t]
\centering
\small
\begin{tabular}{cclc}
\hline
Ranking &Team ID & Team Name          & IDF1  \\ \hline
\textbf{1} & \textbf{6}       & \textbf{UWIPL\_ETRI (ours)} & \textbf{95.36} \\ \hline
2& 9       & HCMIU-CVIP         & 94.17 \\
3& 41      & AILab              & 93.31 \\
4& 51      & FraunhoferIOSB     & 92.84 \\
5& 113     & hust432            & 92.07 \\
6& 133     & ctcore             & 91.09 \\
7& 34      & Team 34            & 91.04 \\
8& 82      & PersonMatching     & 89.81 \\
9& 151     & AIO2022\_VGU       & 89.68 \\
10& 38      & NetsPresso         & 86.76 \\\hline
\end{tabular}
\vspace{-.5em}
\caption{Leaderboard of Track 1 in the AICity Challenge 2023: Multi-Camera People Tracking. Our proposed method obtained an IDF1 score of 95.36 ranking in the first-place.}
\label{table:ranking}
\end{table}

\section{Conclusion}
\label{sec:conclusion}
We proposed a multi-camera people tracking framework that assigns global ID using anchor-based clustering method then calibrates them using spatio-temporal consistency with the self-calibration of cameras. Our approach can successfully improve the accuracy of tracking by identifying key features that are unique to every individual and utilizing the overlap of views between cameras to predict accurate trajectories without needing the actual camera parameters. Experiments and results on a multi-camera dataset with various real and synthetic scenes demonstrated the effectiveness and robustness of our work. Our proposed method ranked first on the public test set of 2023 AI City Challenge Track 1 in IDF1. 

\section{Acknowledgement}
This work was supported by ETRI grant funded by the Korean government (23ZD1120, Regional Industry IT Convergence Technology Development and Support Project). We also want to acknowledge and thank NCHC from Taiwan for providing the computing resources.

{
\bibliographystyle{ieee_fullname}
\bibliography{egbib}
}

\end{document}